%% file: main.tex
\documentclass[letterpaper]{article} 
\usepackage{aaai2026}  
\usepackage{times}  
\usepackage{helvet}  
\usepackage{courier}  
\usepackage[hyphens]{url}  
\usepackage{graphicx} 
\usepackage{natbib}  
\usepackage{caption} 
\usepackage{algorithm}
\usepackage{algorithmic}
\usepackage{chngcntr}
\usepackage{etoolbox}
\usepackage{newfloat}
\usepackage{listings}
\DeclareCaptionStyle{ruled}{labelfont=normalfont,labelsep=colon,strut=off} 
\floatstyle{ruled}
\newfloat{listing}{tb}{lst}{}
\floatname{listing}{Listing}
\usepackage{amsmath}    
\usepackage{amssymb}    
\usepackage{xcolor}     
\usepackage{booktabs}
\usepackage{multirow}
\usepackage{array}

\title{ReCAD: Reinforcement Learning Enhanced Parametric CAD Model Generation with Vision-Language Models}
\author{
    Jiahao Li, Yusheng Luo, Yunzhong Lou, Xiangdong Zhou\thanks{Corresponding author}
}
\affiliations{
    College of Computer Science and Artificial Intelligence, Fudan University, Shanghai, China\\
    Shanghai Key Laboratory of Data Science\\
    \{lijiahao25, ycluo24\}@m.fudan.edu.cn, \{yzlou20, xdzhou\}@fudan.edu.cn
}

\usepackage{bibentry}

\begin{document}

\maketitle
\input{Sections/1_abs}
\input{Sections/2_intro}

\input{Sections/3_related}

\input{Sections/4_method}

\input{Sections/5_exp}

\input{Sections/6_conclusion}

\appendix

\begin{figure*}[t]
\centering
{\huge\textbf{Appendix}}
\end{figure*}

\renewcommand{\thefigure}{A\arabic{figure}}
\renewcommand{\thetable}{A\arabic{table}}
\setcounter{figure}{0}
\setcounter{table}{0}

\input{SupplementaryMaterial/Sections/overview}
\input{SupplementaryMaterial/Sections/data_split}
\input{SupplementaryMaterial/Sections/interface_detail}
\input{SupplementaryMaterial/Sections/details_about_code_format}
\input{SupplementaryMaterial/Sections/prompts_used_in_PLMs}
\input{SupplementaryMaterial/Sections/more_qualit_result}
\input{SupplementaryMaterial/Sections/failure_cases}

\section{Acknowledgments}
The computations in this research were performed using the CFFF platform of Fudan University.

\bibliography{main}

\end{document}

%% file: Sections/1_abs.tex
\begin{abstract}
We present ReCAD, a reinforcement learning (RL) framework that bootstraps pretrained large models (PLMs) to generate precise parametric computer-aided design (CAD) models from multimodal inputs by leveraging their inherent generative capabilities.
With just access to simple functional interfaces (e.g., point coordinates), our approach enables the emergence of complex CAD operations (e.g., \textit{pattern replication} and \textit{mirror}). This stands in contrast to previous methods, which typically rely on knowledge injected through supervised fine-tuning (SFT), offer limited support for editability, and fail to exploit the strong generative priors of PLMs.
Specifically, the ReCAD framework begins by fine-tuning vision-language models (VLMs) to equip them with basic CAD model generation capabilities, where we rewrite  CAD scripts into parameterized code that is leveraged to generate accurate textual descriptions for supervision. 
Then, we propose a novel RL strategy that incorporates parameterized code as guidance to enhance the model’s reasoning on challenging questions. Furthermore, we employ a hierarchical primitive learning process to progressively teach structured and compositional skills under a unified reward function that ensures both geometric accuracy and semantic fidelity.
ReCAD sets a new state-of-the-art in both text-to-CAD and image-to-CAD tasks, significantly improving geometric accuracy across in-distribution and out-of-distribution settings.
In the image-to-CAD task, for instance, it reduces the mean Chamfer Distance from 73.47 to 29.61 (in-distribution) and from 272.06 to 80.23 (out-of-distribution), outperforming existing baselines by a substantial margin.
\end{abstract}

%% file: Sections/2_intro.tex
\section{Introduction}
Prototyping complex computer-aided design (CAD) models is a time-consuming process, as it often involves numerous parts and demands high precision, requiring designers to carefully craft each detail \cite{cherng1998feature, robertson2002cad}. As a result, generative CAD modeling has attracted increasing attention from both the research and industrial communities \cite{daareyni2025generative}. Traditional CAD generation models often employ encoder-decoder architectures \cite{deepcad, hnc-cad}, using modality-specific encoders (e.g., for point clouds \cite{drawstep} or text \cite{text2cad, cad-translator}) and autoregressive decoders to produce CAD models. Recently, CAD generation has shifted towards using pretrained large models (PLMs) to create CAD models, which are represented in the form of code, JSON, or command sequences \cite{cad-gpt, cad-llama, cad-mllm}.

Despite these advancements, the generation of high-precision CAD models from descriptions that encode dimensional and quantitative constraints remains a significant challenge. Ensuring that the generated models align with the specified original design intent is essential for facilitating subsequent reuse and editability \cite{design-intent}. 
Nevertheless, previous methodologies lack the capability to comprehend design intent, as they directly produce CAD sequences consisting solely of low-level parameters (e.g., point coordinates) with modeling commands \cite{cad-llama, iclm-text2cad, flexcad}. The adjustment of these parameters can readily lead to the creation of invalid geometries, such as non-closed loops \cite{cadcrafter}.

At present, the majority of methodologies reconceptualize computer-aided design (CAD) modeling as an endeavor of code generation \cite{cad-llama, cad-coder-rl, cad-coder-sft}. This shift is primarily attributed to the potent capabilities of PLMs \cite{llm-code-proof1, llm-code-proof2}. Nevertheless, PLMs typically act as semantic interpreters \cite{cad-editor} in these works, as their ability to generate and understand CAD models largely relies on the knowledge gained during supervised fine-tuning (SFT). This limits the use of PLMs' generative capabilities, leading to dependence on external knowledge and weak generalization.

Parametric CAD modeling inherently requires precise mathematical reasoning, symbolic manipulation, and the satisfaction of logical constraints \cite{geometry-problem-solve}, which are essential for capturing design intent. Recent advances in reinforcement learning with verifiable rewards (RLVR) \cite{tulu3} have demonstrated promising improvements in the reasoning abilities of PLMs, particularly in mathematics and code generation \cite{deepcoder, deepseekmath}. To this end, we propose ReCAD, a reinforcement learning framework designed to enhance both the reasoning capability and generative performance of PLMs in the context of parametric CAD modeling.

\begin{figure*}
    \centering
    \includegraphics[width=\linewidth]{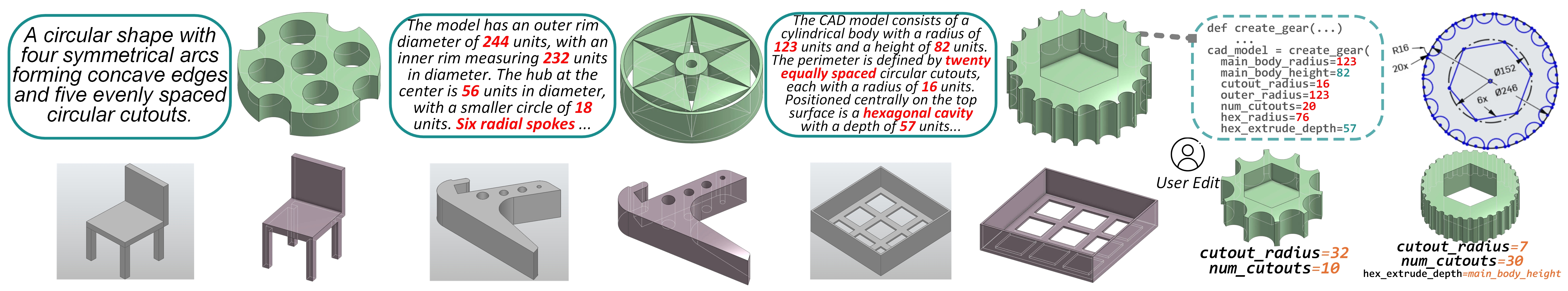}
\caption{\textbf{Visualization of the 3D model generated by our method (ReCAD-VL) based on a single image or textual description (abstract or precise)}. Our method enables the generation of complex and diverse parametric CAD models in code representation, which allows \textbf{precise parameter control} and facilitates easy editing and reuse.}
    \label{fig:framework}
\end{figure*}

Specifically, we first fine-tune vision-language models (VLMs) by converting CAD sequences into parameterized code, which is then used to produce abstract and precise descriptions. These descriptions, together with the corresponding CAD images, are used to perform SFT on both text-to-CAD and image-to-CAD tasks. Building on this, we introduce a reinforcement learning stage using group relative policy optimization (GRPO) \cite{deepseekmath}, leveraging parameterized code as off-policy guidance to enhance the model’s reasoning on inherently challenging questions. Additionally, a hierarchical primitive learning process is proposed to progressively teach compositional skills across abstraction levels. In order to achieve accurate generation, we develop a reward function that assesses both geometric precision and semantic fidelity, thereby ensuring reliable CAD generation.

We show that even with simple function interfaces that expose only the coordinate values of individual curves, while all other components serve merely as hierarchical wrappers, ReCAD enables the emergence of complex CAD operations such as \textit{circular pattern} and \textit{mirror}, which were previously unseen in earlier studies, as illustrated in Figure 1 and Figure 3. 
We also demonstrate that these emergent capabilities substantially enhance the model’s ability to generalize, leading to significantly improved geometric accuracy over the baseline methods under out-of-distribution (OOD) settings.

To summarize, our key contributions are as follows: (1) We introduce ReCAD, a novel reinforcement learning (RL) framework that enables the generation of precise and editable parametric CAD models. (2) We combine supervised fine-tuning with a novel RL strategy in which the parameterized CAD code itself acts as both an off-policy signal and complementary knowledge, boosting reasoning on complex questions. 
(3) We introduce a hierarchical primitive learning process that gradually builds the model’s ability to compose structured designs across abstraction levels under a unified reward for geometric and semantic fidelity.
(4) Experimental results demonstrate that ReCAD outperforms existing methods in both text-to-CAD and image-to-CAD, while exhibiting remarkable generalization capabilities and zero-shot performance.

%% file: Sections/3_related.tex
\section{Related Work}
\textbf{Parametric CAD Sequence Modeling.}
The generation and reconstruction of CAD models have been a long-standing research topic. Along with the emergence of large-scale datasets \cite{deepcad, fusion360}, prior works leverage representation learning methods \cite{deepcad, skexgen, hnc-cad}, where CAD models are first encoded into latent representations, from which the corresponding CAD sequences are decoded. Some methods adopt transformer architectures \cite{signet, text2cad, cad-translator}, encoding inputs such as point clouds, text, or partial CAD models using modality-specific encoders, and then auto-regressively inferring the target CAD sequence. In addition, diffusion-based models \cite{drawstep, raster-sketch, caddreamer, cadcrafter, sketchdnn} have been proposed to generate CAD sequences or directly synthesize B-Rep representations from inputs such as images, point clouds, or partial CAD models. However, these approaches often exhibit limited generalizability and struggle to generate precise CAD models.

\noindent\textbf{PLM for CAD Modeling.}
PLMs have been adapted to CAD-related tasks. To leverage the prior knowledge in PLMs, many works elaborate task-specific datasets and subsequently apply supervised fine-tuning (SFT) to align PLMs with downstream applications involving CAD editing \cite{cad-editor, flexcad, geocad} and generation \cite{cad-gpt, cad-mllm, iclm-text2cad, cadvlm, cad-recode}, where the models are conditioned on various input modalities such as images, text, or point clouds to produce CAD sequences. However, the ability of PLMs to understand and generate CAD sequences remains tightly coupled with the knowledge introduced during the SFT stage, limiting generative priors and hindering faithful modeling of design intent, such as parameter control.
Another line of research explores constrained sketch generation to align outputs with design intent. Vitruvion \cite{vitruvion} and SketchDNN \cite{sketchdnn} generate 2D sketches using token-based and diffusion approaches, respectively, while another work \cite{align-design-intent-iccv} adapts alignment techniques from LLMs. However, these methods are limited to 2D sketches, struggle with complex designs, and rely on external constraint solvers.

\noindent\textbf{Reinforcement Learning for PLM.}
The paradigm of reinforcement learning with verifiable rewards (RLVR) has emerged as a pivotal technique for augmenting the reasoning capabilities of Pre-trained Language Models (PLMs).
Unlike earlier methods based on preferences \cite{rlhf, dpo}, RLVR leverages verifiable signals, yielding strong results on math, code \cite{tulu3, deepseekmath}, and multimodal reasoning benchmarks \cite{vlm-r1, vision-r1}.
On-policy methods like PPO \cite{ppo} and GRPO \cite{deepseekmath} ensure stable training but hinder exploration.
LUFFY \cite{luffy} integrates expert off-policy reasoning into advantage estimation, while another work \cite{learnlikehuman} rewrites expert trajectories using the policy model to reduce distribution mismatch. 
However, obtaining such high-quality Chain-of-Thought (CoT) annotations is both resource-intensive and time-consuming.
Related to our work, CAD-Coder \cite{cad-coder-rl} performs SFT with CoT annotations and RL using expert-level descriptions from the Text2CAD \cite{text2cad} dataset, which contains overparameterized low-level details such as precise coordinates. While this improves CAD generation accuracy, it limits the model's ability to infer low-level attributes from high-level design intent.

%% file: Sections/4_method.tex
\begin{figure*}
    \centering
    \includegraphics[width=\linewidth]{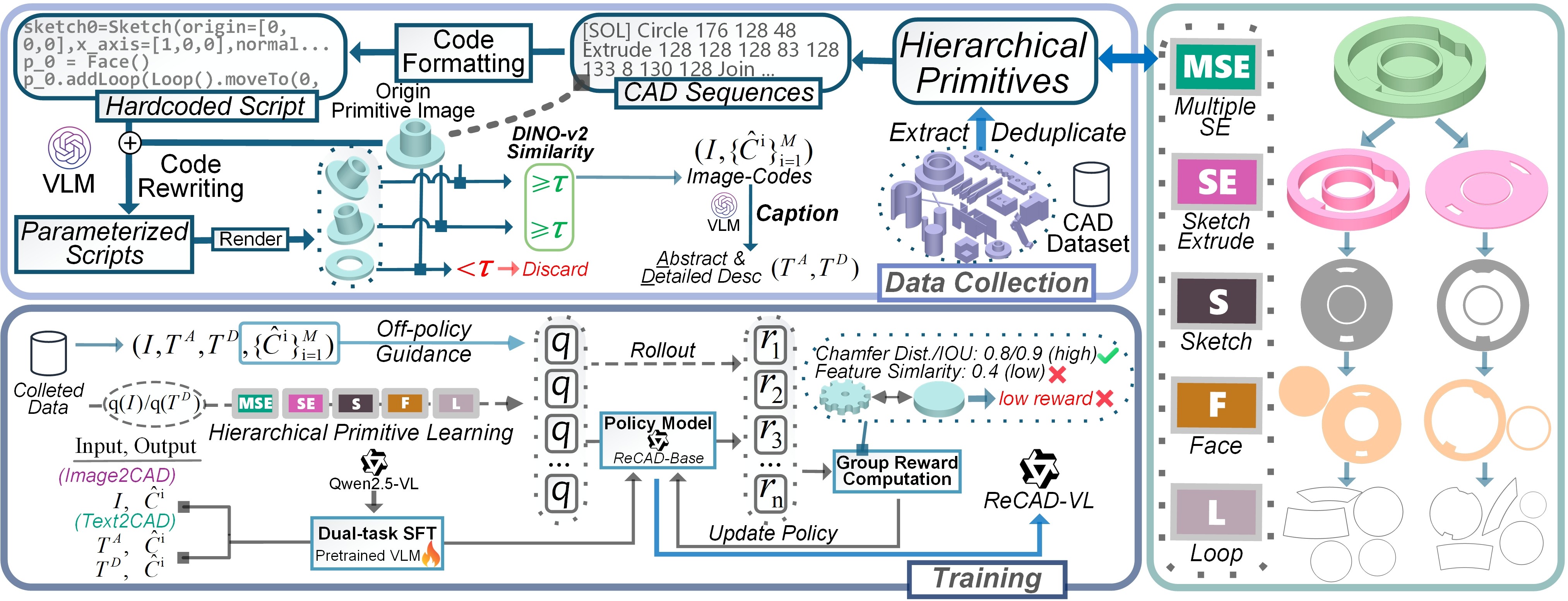}
    \caption{\textbf{Overview of the proposed ReCAD framework.} (1) A CAD model is represented as a primitive hierarchy, where each primitive is converted into parameterized code, which is further leveraged to generate precise textual descriptions. (2) We first perform supervised fine-tuning (SFT) for basic CAD code generation, then optimize the model via reinforcement learning guided by parameterized code, with reward functions enforcing both geometric fidelity and feature-level consistency.}
    \label{fig:framework}
\end{figure*}

\section{Method}
\label{sec:method}
We present ReCAD, a framework that leverages VLMs and RL to generate precise parametric CAD models from a single image or text. We first introduce the problem and notation, then describe a two-stage training process: supervised fine-tuning (SFT) and reinforcement learning (RL).

\subsection{Problem Definition and Notation}\label{sec:problem-definition}
ReCAD takes as input a CAD image $\mathit{I}$ or a textual description $T$, either detailed ($T^D$) or abstract ($T^A$), and translates it into parametric CAD code $\hat C$ representing the final model.

\textbf{Sketch-Extrude Paradigm.}
We build upon the widely used \textit{sketch-extrude (SE)} design paradigm, where a CAD model $D$ is composed of multiple SE pairs, denoted as \textit{MSE}. Each sketch \textit{S} contains multiple faces; each face \textit{F} consists of one or more loops \textit{L}; and each loop is a closed path formed by one or more curves. Based on this hierarchical structure, we define five levels of hierarchical primitives:
\begin{equation}
\label{eq-hierarchy}
\mathcal{P} = \{\textit{L}, \textit{F}, \textit{S}, \textit{SE}, \textit{MSE}\},
\end{equation}
where $P \in \mathcal{P}$ denotes any primitive in the hierarchy. A complete CAD model can be represented by a primitive $P$ when $P = \textit{SE}$ or $P = \textit{MSE}$.

\textbf{CAD Sequence Representation.}
Instead of adopting original CAD sequence representations (e.g., \texttt{Line 128 128}) or translating CAD sequences into CADQuery \cite{cadquery} code, we introduce a lightweight set of function interfaces tailored to $\mathcal{P}$. At the lowest level, curves are defined via coordinate-based interfaces, and higher-level primitives $P$ are progressively wrapped to ultimately form an \textit{MSE} object that represents a complete CAD model. 
Details of the interface design are provided in the Appendix.
We define a function $f()$ that converts a primitive $P$ into our interface-based CAD code $C$, as follows:
\begin{equation}
C = f(P).
\end{equation}

\subsection{Supervised Fine-Tuning on Parameterized Code}\label{sec:sft-stage}
Although reinforcement learning with verifiable rewards (RLVR) has shown promising results, it remains insufficient for encouraging capabilities beyond the inherent limitations of the base model. Due to the limited proficiency of existing VLMs in CAD-specific tasks \cite{vlm-struggle-w-visual}, we first employ a supervised fine-tuning (SFT) stage to establish foundational capabilities in parametric CAD code generation.

\textbf{Parameterized Code.}
Prior work \cite{cad-llama} has demonstrated that converting CAD sequences into code representations enhances model performance in CAD generation. However, such naive conversions fail to support modular coding practices and often produce unflexible, overfit-prone representations (referred to as hardcoded CAD code), which may limit their ability to generalize to out-of-distribution (OOD) samples.

To this end, we rewrite hardcoded CAD code into parameterized code using a VLM, aiming to unleash the LLM’s strong prior in programmatic code generation and to provide a more flexible representation of CAD code.
For a primitive \( P \), we first obtain its hardcoded code \( C = f(P) \) and render a canonical-view image \( I \) of \( P \). Specifically, we render a canonical view if \( P \in \{\textit{SE}, \textit{MSE}\} \), and a top-down view otherwise. The image-code pair \( (I, C) \) is then fed into the VLM to sample \( N \) candidate parameterized codes:
\begin{equation}
\{\hat{C}^i\}_{i=1}^{N} = \text{VLM}(I, C).
\end{equation}
To select high-quality candidates, we render each $\hat{C}^i$ into an image $\hat{I}_i$ and use a pretrained DINOv2 \cite{dinov2} model as an encoder $E(\cdot)$ to compute the cosine similarity between $E(\hat{I}_i)$ and $E(I)$. We retain only candidates who satisfy:
\begin{equation}
\mathcal{C} = \left\{ \hat{C}^i \;\middle|\; \cos(E(\hat{I}_i), E(I)) > \tau_s \right\},
\end{equation}
where $\tau_s$ is a similarity threshold. If no valid candidates exist, we revert to the original code: $\mathcal{C} = \{C\}$

\textbf{Text Description Generation.}
To generate textual descriptions for CAD models, existing methods often rely on VLMs and complex annotation pipelines. However, these methods typically struggle to capture precise information (e.g., scale, quantity), and the resulting descriptions, even when incorporating CAD sequence information, tend to be overly verbose with redundant parameter details \cite{text2cad}; as such, low-level information is difficult for VLMs to interpret.

Instead, we leverage the parameterized code $\mathcal{C}$, which inherently contains semantically meaningful and precise information about both scale and quantity. Such information is explicitly encoded and is thus easily interpretable by VLMs. By feeding both the parameterized code and image $I$ into a VLM, we obtain textual descriptions:
\begin{equation}
T^A, T^D = \text{VLM}(I, \hat{C}^i),
\end{equation}
where $T^A$ and $T^D$ denote abstract and detailed descriptions, respectively, and $\hat{C}^i \in \mathcal{C}$.

\begin{figure*}[t]
    \centering
    \includegraphics[width=\linewidth]{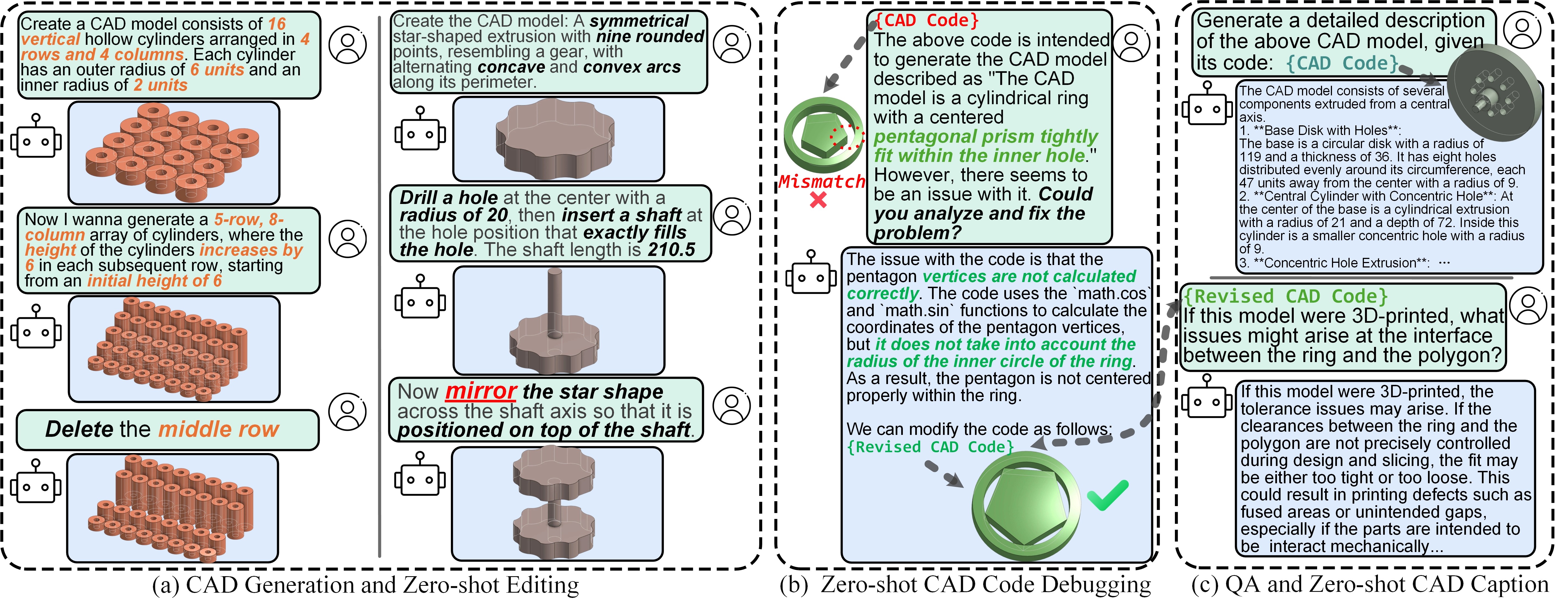}
    \caption{\textbf{Dialog results on CAD generation and zero-shot CAD-related tasks.} Despite being trained exclusively on CAD generation tasks, ReCAD-VL exhibits impressive zero-shot performance across multiple CAD-related tasks, including understanding, editing, debugging, and descriptive captioning. 
    More detailed qualitative results are provided in the Appendix.
    }
    \label{fig:editing}
\end{figure*}

\textbf{Supervised Fine-Tuning.}
We employ the standard causal language modeling (CLM) objective to continuously fine-tune the LLM on two tasks: text-to-CAD and image-to-CAD. Formally, given the dataset \(\mathcal{D}_{\text{SFT}} = \{q_i, \mathcal{C}_i\}_{i=1}^N\), where \(q_i = T^D_i / T^A_i\) is for the text-to-CAD task and \(q_i = I_i\) is for the image-to-CAD task, we optimize the LLM via maximum likelihood:
\begin{equation}
\mathcal{L}(\mathcal{D}_{\text{SFT}}) = - \sum_{i=1}^{N} \log \pi_{\theta}(\mathcal{Y}_i \in \mathcal{C}_i \mid \mathcal{X}_i = q_i).
\end{equation}
Here, \(\pi_{\theta}\) denotes the LLM with trainable parameters $\theta$, and \(\mathcal{X}_i\), \(\mathcal{Y}_i\) represent the input and output, respectively. 

To retain the LLM’s general abilities, we augment the training corpus with a small proportion of data sampled from \textit{UltraChat} and \textit{OpenCodeReasoning}. 
After fine-tuning, the model acquires basic CAD code generation abilities, and we name this model ReCAD-Base.

\subsection{Reinforcement Learning Under Guidance}\label{sec:rl-guidance}
We employ RL after SFT to enhance the model’s generalization ability and its capability to generate precise geometry. Similar to the SFT stage, we perform RL on both text-to-CAD and image-to-CAD tasks and refer to the trained model as ReCAD-VL. In contrast to the SFT stage, the text-to-CAD input in the RL stage includes only the precise description $T^D$, since one abstract description $T^A$ may map to multiple valid geometries, complicating reward design. We employ group relative policy optimization (GRPO), with ReCAD-Base serving as the base model. The GRPO objective is defined as:
\begingroup
\small
\begin{equation}
\mathcal{J}(\theta) =
 \frac{1}{N}\sum_{i=1}^N\frac{1}{|\tau_i|} \sum_{t=1}^{|\tau_i|}\text{CLIP}(r_{i,t}(\theta, q),A_{i},\epsilon) - \beta \mathbb{D}_{\text{KL}}[\pi_{\theta} || \pi_{\text{ref}}],
\end{equation}
\endgroup
where each question $q$ is associated with a set of sampled solutions $\{\tau_i\}_{i=1}^{N}$ generated from $\pi_{\theta_{\text{old}}}$. The advantage $A_i$ is computed using normalized rewards within the group:
\begin{equation}
A_i = \frac{R(\tau_i) - \mathrm{mean}(\{R(\tau_j)\}_{j=1}^{N})}{\mathrm{std}(\{R(\tau_j)\}_{j=1}^{N})},
\end{equation}
where $R(\cdot)$ is the reward function. The clipped surrogate objective $\text{CLIP}(r_{i,t}(\theta, q), A_i, \epsilon)$ is defined as:
\begin{equation}
\text{CLIP}(r, A, \epsilon) = \min \left[ r \cdot A, \; \mathrm{clip}(r, 1 - \epsilon, 1 + \epsilon) \cdot A \right],
\end{equation}
where $r_{i,t}(\theta, q) = \pi_\theta(\tau_{i,t} \mid q, \tau_{i,<t}) / \pi_{\theta_{\text{old}}}(\tau_{i,t} \mid q, \tau_{i,<t})$ denotes the importance sampling ratio.

\textbf{Learn Under Guidance.}\label{sec:curriculum}
Despite the success of RLVR, its on-policy nature restricts learning to self-generated outputs and fundamentally limits it to the base LLM. Inspired by the strength of LLMs in few-shot in-context learning \cite{icl},
we treat the rewritten parameterized codes $\mathcal{C}$ as off-policy guidance during the rollout, providing complementary knowledge that helps the model develop coherent reasoning guided by correct code. The objective of this process is defined as follows:
\begingroup
\small
\begin{align}
&\mathcal{\hat J}(\theta; \mathcal{C}) = \frac{1}{N-|\mathcal{C}|}\sum_{i=1}^{N-|\mathcal{C}|}\frac{1}{|\tau_i|} \sum_{t=1}^{|\tau_i|}\text{CLIP}(r_{i,t}(\theta,q),A_{i},\epsilon)\notag \\
& + \frac{1}{|\mathcal{C}|}\sum_{j=1}^{|\mathcal{C}|}\frac{1}{|\tau_j|} \sum_{t=1}^{|\tau_j|}\text{CLIP}(\hat r_{j,t}(\theta, q, C_j),A_{j},\epsilon)\notag \\
&- \beta \mathbb{D}_{\text{KL}}[\pi_{\theta} || \pi_{\text{ref}}],
\end{align}
\endgroup
where $C_j \in \mathcal{C}$ denotes the $j$-th parameterized code used for guidance, and $\tau_j$ is the corresponding solution generated by $\pi_{\theta_{\text{old}}}$ conditioned on the question $q$ and $C_j$. The importance sampling ratio is defined as $\hat r_{j,t}(\theta, q, C_j) = \pi_\theta(\tau_{j,t} \mid q, \tau_{j,<t}) / \pi_{\theta_{\text{old}}}(\tau_{j,t} \mid q, C_j, \tau_{j,<t})$. The advantage values $A$ are computed as in Equ. (8) over the group of solutions generated with and without guidance, i.e., $\tau_j$ and $\tau_i$, respectively.

We define the RL training dataset as $\mathcal{D}_{\mathrm{RL}} = \{q_i, \mathcal{C}_i\}_{i=1}^N$. 
Before RL training, we identify difficult questions by prompting the model with each \( q_i \), sampling \( N \) solutions, and computing the maximum reward \( \max\{R(q_i)\} \).
If the reward falls below a predefined threshold $\tau_h$, we identify $q_i$ as a hard question and use the guidance-based objective $\hat{\mathcal{J}}(\theta; \mathcal{C}_i)$. Otherwise, we apply the standard GRPO objective $\mathcal{J}(\theta)$. Thus, the final training objective is:
\begingroup
\begin{equation}
\begin{aligned}
\mathcal{L}_{\mathrm{RL}}(\theta) &=\mathbb{E}_{(q_i, \mathcal{C}_i) \sim \mathcal{D}_{\mathrm{RL}}} \;  \\
 & \left[
\mathbf{1}_{\mathrm{hard}}(q_i) \cdot \hat{\mathcal{J}}(\theta; \mathcal{C}_i) \right.
 \left. + \left(1 - \mathbf{1}_{\mathrm{hard}}(q_i)\right) \cdot \mathcal{J}(\theta) \right],
\end{aligned}
\end{equation}
\endgroup
where $\mathbf{1}_{\mathrm{hard}}(q_i)$ is an indicator function that evaluates to $1$ if the maximum reward for question $q_i$ falls below a predefined threshold $\tau_h$, and $0$ otherwise:
\begin{equation}
\mathbf{1}_{\mathrm{hard}}(q_i) =
\begin{cases}
1 & \text{if } \max\{R(q_i)\} < \tau_h \\
0 & \text{otherwise}
\end{cases}.
\end{equation}

\textbf{Hierarchical Primitive Learning.}\label{sec:curriculum} 
We introduce hierarchical primitive learning (HPL) as a curriculum learning strategy tailored to the CAD generation task. Curriculum learning \cite{cl0, cl1} is a widely adopted strategy in complex task domains, wherein the learning process is organized into a sequence of increasingly difficult examples. This paradigm mirrors the human learning process \cite{cl-cognition}, enabling models to progressively acquire fundamental skills before tackling more intricate challenges. In our setting, CAD models follow the sketch-extrude (SE) paradigm, which naturally forms a hierarchical structure from curves to loops, faces, sketches, and extrusions.
Based on the defined hierarchical primitives $\mathcal{P}$ (Equ. (1)), we design a curriculum that follows the inherent structure of CAD models, progressing through \textit{L}, \textit{F}, \textit{S}, \textit{SE}, and \textit{MSE}. Each stage introduces increasing complexity and builds upon the structure of the preceding one, allowing the model to acquire compositional skills in a gradual and structured manner. For each level, we further organize training samples by difficulty, which is defined by the number of curves involved. The model learns in a simple-to-complex order, starting with primitives composed of fewer curves and gradually moving to those with richer geometric structures.

\begin{table*}
\centering
\setlength{\tabcolsep}{3pt}
\begin{tabular}{lccccccccc}
\toprule[1.5pt]
\multirow{2}{*}{Methods} 
& \multicolumn{4}{c}{\textbf{DeepCAD}} 
&
& \multicolumn{4}{c}{\textbf{Fusion 360 (Out-of-Distribution)}} 
\\
\cmidrule(lr){2-5} \cmidrule(lr){7-10}
& \textit{P}-F1$\uparrow$
& \textit{Median} CD$\downarrow$ & \textit{Mean} CD$\downarrow$ & \multicolumn{1}{c}{\multirow{1}{*}{IR$\downarrow$}}
&
& \textit{P}-F1$\uparrow$
& \textit{Median} CD$\downarrow$ & \textit{Mean} CD$\downarrow$ & \multicolumn{1}{c}{\multirow{1}{*}{IR$\downarrow$}}
\\ \midrule
GPT-4o
& $50.55$ & $107.55$ & $165.67$ & $15.14$ &
& $55.10$ & $125.68$ & $187.90$ & $19.81$
\\
DeepSeek-V3
& $50.41$ & $110.67$ & $167.90$ & $15.43$ &
& $\underline{54.34}$ & $135.10$ & $190.27$ & $18.31$
\\
Qwen3-235B-A22B
& $52.45$ & $109.51$ & $157.49$ & $24.84$ &
& $53.07$ & $121.83$ & $182.74$ & $26.55$
\\ \midrule
CAD-Translator \cite{cad-translator}
& $51.28$ & $92.70$ & $162.47$ & $3.87$ &
& $45.55$ & $192.51$ & $243.59$ & $3.86$
\\
Text2CAD \cite{text2cad}
& $52.35$ & $91.26$ & $160.49$ & $2.75$ &
& $45.27$ & $254.95$ & $285.95$ & $3.93$
\\
CAD-LLaMA \cite{cad-llama}
& $\underline{60.02}$ & $\underline{41.77}$ & $\underline{98.12}$ & $\mathbf{0.39}$ &
& $50.47$ & $\underline{60.36}$ & $\underline{142.48}$ & $\underline{1.29}$
\\ \midrule
ReCAD-VL (Ours) 
& $\mathbf{61.48}$ & $\mathbf{34.31}$ & $\mathbf{72.47}$ & $\underline{0.81}$ &
& $\mathbf{55.25}$ & $\mathbf{34.67}$ & $\mathbf{84.89}$ & $\mathbf{0.93}$
\\
\bottomrule[1.5pt]
\end{tabular}
\caption{Text-to-CAD generation results on DeepCAD (left) and Fusion 360 (right). We report average F1 score for primitives (\textit{P}-F1), mean and median CD and IR. CD is multiplied by $10^3$. \textbf{Bold} and \underline{underline} indicate the best and the second best result.}
\label{tab:text2cad1}
\end{table*}

\textbf{Reward Design.}\label{sec:gsr}
To ensure that the generated CAD models are both geometrically accurate and visually faithful, we design a reward function that combines intersection-over-union under optimal alignment \cite{cad-coder-sft}, denoted as $\text{IOU}_{best}$, with feature-level similarity:
\begin{align}
R(y_\pi, \Omega)&=\lambda_1\cdot \text{min}\{\text{IOU}_{best}(\hat \Omega, \Omega),\ \phi(\text{sim}(\hat I, I),\ \tau)\} \notag \\
& + \lambda_2\cdot R_{f}(y_\pi),
\end{align}
where \( y_\pi \) denotes the model-generated output sequence. The geometric consistency between the predicted geometry \( \hat\Omega \) and ground-truth \( \Omega \) is measured by \( \text{IOU}_{best}(\hat\Omega, \Omega) \), which computes the IOU and selects the alignment with maximal overlap. The image similarity \( \text{sim}(\hat I, I) = \text{cos}(E(\hat I), E(I)) \) is the cosine similarity between the rendered images of \( \hat\Omega \) and \( \Omega \). The function \( \phi(s, \tau) = \max(0,\ (s - \tau)/(1 - \tau)) \) applies thresholded linear scaling, with \(\tau\) set to 0.55 in our experiments. Scalars \( \lambda_1 \) and \( \lambda_2 \) are weighting coefficients. The formatting reward \( R_f(y_\pi) \) equals 1 if \( y_\pi \) begins with a valid \texttt{<think>...</think>} block and 0 otherwise.

For image-to-CAD, the input images do not contain absolute scale information. Therefore, we normalize the solid geometry before computing the reward:
\begin{equation}
    n(\Omega)=\{\frac{\mathbf{x}-\bar{\mathbf{x}}}{\sqrt{\frac{\operatorname{tr}(\mathbf{I})} {2\times \mathrm{V o l} ( \Omega_{2} )}}}\mid\mathbf{x}\in \Omega\},
\end{equation}
where $\mathbf{I}$ is the inertia matrix and $\bar{\mathbf{x}}$ is the centroid of $\Omega$.

%% file: Sections/5_exp.tex
\section{Experiment}
We present the experimental details and evaluate our method on text-to-CAD and image-to-CAD generation tasks, comparing it with SOTA methods to demonstrate its effectiveness.

\subsection{Experimental Setups}
\textbf{Datasets.}
Our model, trained on the DeepCAD training set, is evaluated on the DeepCAD and Fusion360 Gallery datasets. After filtering out trivial models, such as cylinders and cubes, we have approximately 90,000 models from the DeepCAD dataset. We extract hierarchical primitives ${P} \in \mathcal{P}$ from the training set, render images, and calculate pairwise similarities using DINOv2 embeddings. Primitives with similarity scores above 0.95 are deemed duplicates and removed. For fair evaluation, we use Text2CAD \cite{text2cad} intermediate level descriptions for DeepCAD's test set and generate descriptions for Fusion360 as per CAD-LLaMA \cite{cad-llama}.
\newline\textbf{Implementation Details.}
We select Qwen2.5-VL-7B-Instruct \cite{qwen2-5-vl} as the base model for SFT. The learning rate is set to $1.0 \times 10^{-5}$ for supervised fine-tuning (SFT) with a context length of 3072, and $2.0 \times 10^{-6}$ for reinforcement learning (RL) with a context length of 8192, with cosine learning rate scheduling. We employ GPT-4o \cite{gpt-4o} for both code rewriting and text description generation. Filtering thresholds are set as $\tau_s = 0.95$ and $\tau_h = 0.8$. The weighting coefficients $\lambda_1$ and $\lambda_2$ are empirically set to $0.1$ and $0.9$, respectively. 
We adopt DeepSpeed \cite{deepspeed} and Flash-Attention \cite{flashattention} to accelerate training.
Our framework is built upon veRL \cite{verl}, where we sample 8 candidate solutions per question $q$ during rollout, with all experiments run on 8 NVIDIA A800 80GB GPUs.
\begin{figure}[h]
    \centering
    \includegraphics[width=\columnwidth]{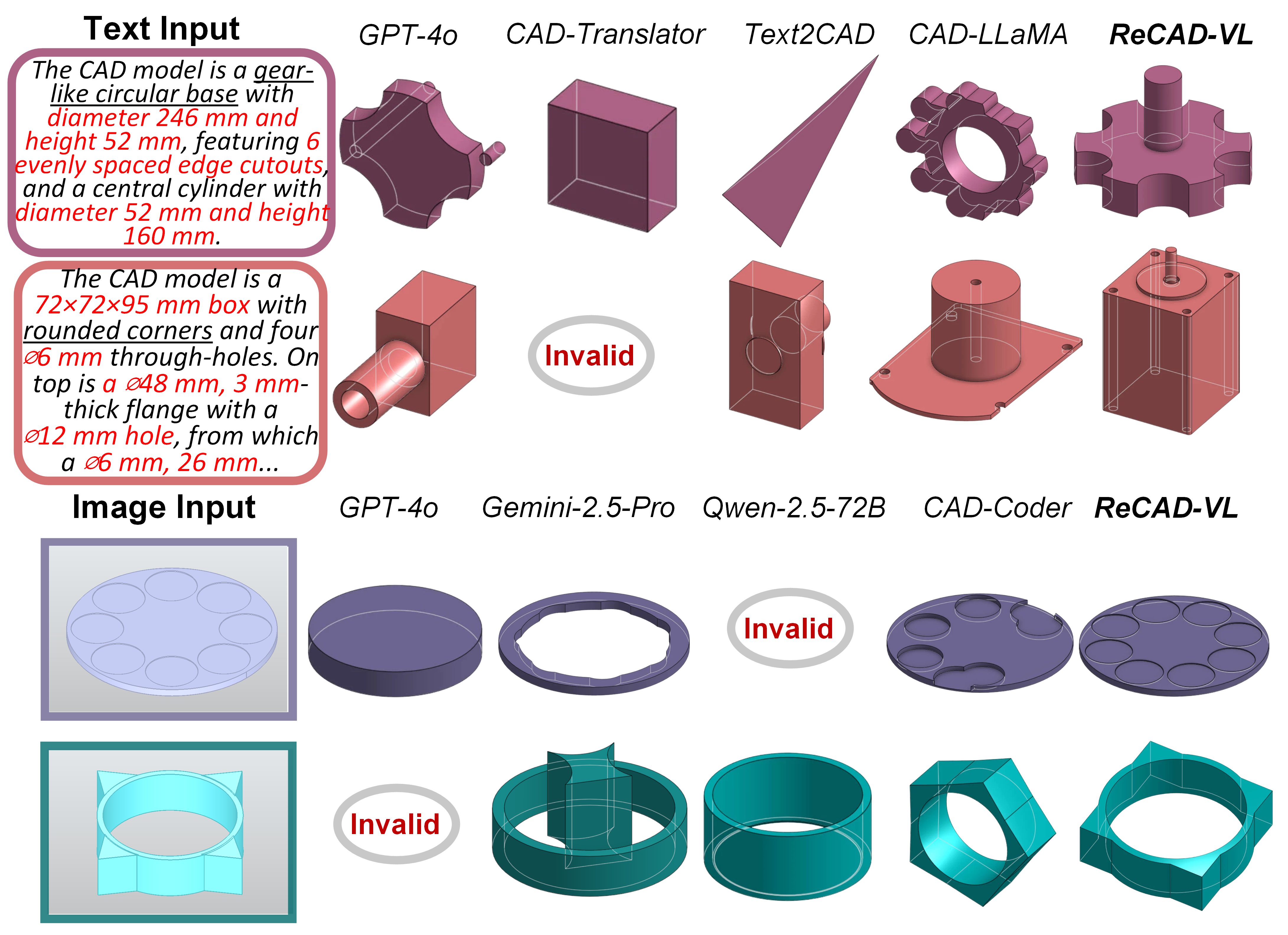}
    \caption{Qualitative comparison of different methods on text-to-CAD and image-to-CAD tasks.}
    \label{fig:visual_compare}
\end{figure}
\newline\textbf{Metrics.}
We use the Chamfer Distance (CD) to evaluate the geometric alignment between the predicted and ground-truth CAD models, and the Invalidity Ratio (IR) to measure the proportion of invalid CAD models across all tasks. In addition, for text-to-CAD generation, we adopt the average F1 score of primitives (\textit{P}-F1) \cite{text2cad}, while for image-to-CAD generation, we use $\text{IOU}_{best}$ from CAD-Coder \cite{cad-coder-sft}, which measures the intersection-over-union under the best alignment between the generated and ground-truth CAD models.

\begin{table*}
\centering
\setlength{\tabcolsep}{3pt}
\begin{tabular}{lcccc cccc}
\toprule[1.5pt]
\multirow{2}{*}{Methods} 
& \multicolumn{4}{c}{\textbf{DeepCAD}} 
& \multicolumn{4}{c}{\textbf{Fusion 360 (Out-of-Distribution)}} 
\\
\cmidrule(lr){2-5} \cmidrule(lr){6-9}
& IOU$_{best}$$\uparrow$
& \textit{Median} CD$\downarrow$ & \textit{Mean} CD$\downarrow$ & IR$\downarrow$
& IOU$_{best}$$\uparrow$
& \textit{Median} CD$\downarrow$ & \textit{Mean} CD$\downarrow$ & IR$\downarrow$
\\ \midrule
GPT-4o
& $37.88$ & $198.34$ & $274.23$ & $11.51$
& $41.75$ & $184.45$ & $308.14$ & $8.72$
\\
Gemini-2.5-Pro
& $45.19$ & $160.26$ & $219.98$ & $37.38$
& $44.95$ & $158.46$ & $326.31$ & $39.36$
\\
Qwen-2.5-72B
& $34.51$ & $180.12$ & $258.02$ & $5.01$
& $31.19$ & $175.27$ & $313.75$ & $9.41$
\\ \midrule
CAD-Coder \cite{cad-coder-sft}
& $\underline{61.23}$ & $\underline{8.09}$ & $\underline{73.47}$ & $\mathbf{1.05}$
& $\underline{45.32}$ & $\underline{84.02}$ & $\underline{272.06}$ & $\underline{2.23}$
\\ \midrule
ReCAD-VL (Ours) 
& $\mathbf{63.14}$ & $\mathbf{7.45}$ & $\mathbf{29.61}$ & $\underline{1.12}$
& $\mathbf{54.93}$ & $\mathbf{17.01}$ & $\mathbf{80.23}$ & $\mathbf{0.91}$
\\
\bottomrule[1.5pt]
\end{tabular}
\caption{Image-to-CAD generation results on DeepCAD (left) and Fusion 360 (right). We report $\text{IOU}_{best}$, mean and median CD and IR. CD is multiplied by $10^3$. \textbf{Bold} and \underline{underline} indicate the best and the second best result.}
\label{tab:text2cad}
\end{table*}

\subsection{Results on Text-to-CAD Generation}
In the text-to-CAD task, we provide the functional interfaces for strong PLMs such as DeepSeek-V3 and GPT-4o, and employ a two-shot approach, with detailed prompts provided in the Appendix. As shown in Table 1, ReCAD-VL achieves more accurate geometry generation compared to previous SOTA methods. On the in-distribution setting, our method outperforms CAD-LLaMA by $\sim$7 and 26 in terms of median and mean Chamfer Distance (CD), respectively. This indicates that our approach produces more precise parameters, thereby generating higher-fidelity models and significantly surpassing other baseline methods, as illustrated in Figure 4. These results suggest that both previous methods and PLMs, despite being able to infer correct command types (as reflected by \textit{P}-F1), struggle to predict accurate parameters.

In the Out-of-distribution setting, our method significantly outperforms all baselines, surpassing CAD-LLaMA by $\sim$26 and 58 in median and mean CD, respectively. The comparable performance of our model on both in-distribution and out-of-distribution data highlights its strong generalization ability. Due to the lack of relevant prior knowledge and the absence of training, PLMs, while maintaining consistent performance across distributions, still struggle to generate precise models. In contrast, previous methods show substantial performance drops on out-of-distribution data, indicating potential overfitting to in-distribution data. This may elucidate their higher invalidity ratio (IR) observed within the Fusion 360 dataset, as exemplified by CAD-LLaMA, which conversely achieves marginally superior results compared to our method in the in-distribution context. Notably, without task-specific training, our model shows impressive performance on CAD-related tasks such as editing and debugging (Figure 3), demonstrating the effectiveness of leveraging PLMs’ intrinsic code generation and reasoning abilities.

\subsection{Results on Image-to-CAD Generation}
In the image-to-CAD task, we use a single image as input and prompt the baseline PLMs using a method similar to that used in the text-to-CAD task. As shown in Table 2, our model outperforms previous methods on nearly all metrics. In particular, we achieve a significant lead over CAD-Coder \cite{cad-coder-sft} 44 in terms of the mean Chamfer Distance (CD), indicating that our approach is more robust and consistently produces high-fidelity results across different input images.
In the Out-of-distribution setting, our model also exhibits strong robustness and generalization ability, outperforming CAD-Coder \cite{cad-coder-sft} and PLMs by a large margin across all metrics. CAD-Coder \cite{cad-coder-sft}, which is similar to \cite{cad-coder-rl}, converts the code into the CADQuery format. However, these methods neither leverage the pre-existing knowledge and generative capabilities of PLMs nor exploit the inherent strengths of CADQuery itself. As a result, similar to the text-to-CAD task, their performance drops significantly on out-of-distribution data.

\begin{table}[ht]
\centering
\setlength{\tabcolsep}{3pt}
\begin{tabular}{lcccc}
\toprule[1.5pt]
Methods
& \textit{P}-F1
& \textit{Median} CD$\downarrow$ & \textit{Mean} CD$\downarrow$ & IR$\downarrow$
\\ \midrule
SFT \textit{only}
& $53.53$ & $84.78$ & $155.67$ & $3.21$
\\ 
RL \textit{only}
& $55.61$ & $107.32$ & $179.50$ & $4.77$
\\
\textit{w/o HPL}
& $59.63$ & $44.64$ & $90.83$ & $2.42$
\\
\textit{w/o Guidance}
& $60.03$ & $42.85$ & $87.34$ & $0.93$
\\
Ours
& $61.48$ & $34.31$ & $72.47$ & $0.81$
\\
\bottomrule[1.5pt]
\end{tabular}
\caption{Ablation experiment on the DeepCAD dataset for the text-to-CAD task.}
\label{tab:ablation}
\end{table}

\subsection{Ablation Study} 
We conducted ablation studies on the text-to-CAD task to evaluate the effectiveness of different training strategies in CAD generation. As shown in Table 3, using SFT alone or RL alone results in suboptimal performance. SFT primarily relies on memory-based imitation, which lacks sufficient generalization and often requires repeated sampling to achieve desirable outputs. RL, on the other hand, suffers from low learning efficiency and is limited by the model's intrinsic capabilities. 
This suggests that neither method alone is adequate for the CAD domain, particularly in the absence of prior task-specific knowledge. 
Combining SFT and RL helps mitigate these limitations: SFT introduces external knowledge that the model cannot acquire autonomously, while RL further enhances the model’s ability to generalize by reinforcing the knowledge learned during SFT.
Moreover, providing guidance as complementary knowledge enables compensation for the limitations of reinforcement learning (RL) while leveraging the model’s in-context learning capability, thereby further enhancing the generation quality. Without HPL, reconstruction errors and failure rates increase, indicating that it enables gradual compositional learning and yields more reliable CAD structures.

%% file: Sections/6_conclusion.tex
\section{Conclusion}
We present ReCAD, a reinforcement learning framework leveraging PLMs to generate precise parametric CAD models. Our method consists of two stages: initially, a supervised fine-tuning phase is employed to transform hardcoded Computer-Aided Design (CAD) scripts into parameterized code, thereby facilitating the generation of precise descriptions. Subsequently, a reinforcement learning phase is implemented, utilizing the parameterized code as a guiding framework. This phase incorporates a hierarchical primitive learning paradigm aimed at developing compositional modeling competencies under a unified reward structure, thus ensuring both geometric and semantic fidelity. Experimental results show that ReCAD performs strongly on in-distribution data and generalizes well to out-of-distribution cases. This demonstrates that by exploiting pre-existing knowledge within PLMs, ReCAD enables precise parameter control while preserving design intent in the generation of CAD models, making it well suited for real-world CAD applications.

%% file: SupplementaryMaterial/Sections/overview.tex
\section{Overview}
This supplementary material provides extended information to complement the main paper, including the following:

\begin{enumerate}
    \item Statistical information about the dataset used during the supervised fine-tuning (SFT) stage, including the proportion of different data sources;
    \item The design of our CAD interface;
    \item Detailed information on our CAD representation, including parameter design and quantization strategies;
    \item Implementation details of the baseline methods, including the prompts used;
    \item Additional qualitative results, including full model inputs and outputs, as well as more generation and zero-shot editing examples;
    \item Failure cases.
\end{enumerate}

%% file: SupplementaryMaterial/Sections/data_split.tex
\section{Dataset Statistics}
During the supervised fine-tuning (SFT) stage, we constructed a composite training dataset comprising our curated in-house data, selected portions of the \textit{UltraChat} dataset \cite{ultrachat}, a general-purpose dialogue corpus, and a small subset of the \textit{OpenCodeReasoning} dataset \cite{opencodereasoning}, which helps maintain the general code understanding capabilities of pretrained large models (PLMs). Table 1 summarizes the data sources used in SFT, including the number of samples drawn from each and their respective proportions in the final training mixture.

\begin{table}[h]
\centering
\setlength{\tabcolsep}{3pt}
\begin{tabular}{l|cc}
\toprule[1.5pt]
Data Source
& Items & Proportion
\\ \midrule
\textit{UltraChat}
& $85k$ & $23\%$
\\
\textit{OpenCodeReasoning}
& $20k$ & $5\%$
\\
Ours
& $254k$ & $72\%$
\\ 
\bottomrule[1.5pt]
\end{tabular}
\caption{Composition of the supervised fine-tuning (SFT) dataset, including the number of samples and their proportions from each data source.}
\label{tab:sft_data_statistics}
\end{table}

%% file: SupplementaryMaterial/Sections/interface_detail.tex
\section{CAD Interface Design}
We define a simple interface to represent CAD models in code, as illustrated in Listing 1. At the lowest level, we provide access to coordinate values for each type of curve, while higher-level components are organized hierarchically through structured encapsulation. The complete CAD model is represented by a \texttt{CADModel} class, which encapsulates multiple \textit{sketch-extrude} (SE) pairs. Each \texttt{Sketch} contains one or more \texttt{Face} objects, and each \texttt{Face} comprises one or more \texttt{Loop} objects that ultimately provide the concrete curve coordinate information.

Although CADQuery provides a similar interface for representing CAD models, such conversions may introduce syntactic errors that require post-processing~\cite{cadquery-counter-example1, cadquery-counter-example2}. Moreover, we note that this conversion is merely an engineering optimization rather than a methodological advance. Instead, our research focuses on effectively leveraging reinforcement learning (RL) and the strong generative priors of pretrained language models (PLMs) to generate parameterized CAD models, without being limited to specific code representations. Experimental results clearly demonstrate the effectiveness of our method, and we believe it remains effective even after conversion into CADQuery.

\begin{figure*}[!htbp]
\centering
\begin{minipage}{0.95\textwidth}
\begin{lstlisting}[language=Python, caption={\textbf{Interface definition for representing CAD models in code.} The interface organizes CAD geometry hierarchically, from low-level curve coordinates up to high-level \texttt{CADModel} structures. Each \texttt{CADModel} encapsulates multiple sketch-extrude (SE) pairs, composed of \texttt{Sketch}, \texttt{Face}, and \texttt{Loop} components.}, captionpos=t]
from typing import Tuple, Union, Self

class CADModel:
    def __init__(self) -> None: ...
    def addSE(self, sketch: Sketch, extrude: Extrude, boolean_op: str) -> None: ...

class Extrude:
    def __init__(self, distance: Union[float, Tuple[float, float]]) -> None: ...

class Sketch:
    def __init__(self, origin: list[float], x_axis: list[float], normal: list[float]) -> None: ...
    def addFace(self, *faces: Face) -> None: ...

class Face:
    def __init__(self) -> None: ...
    def addLoop(self, *loops: Loop) -> None: ...

class Loop:
    def __init__(self) -> None: ...
    def moveTo(self, x: float, y: float) -> Self: ...
    def lineTo(self, x: float, y: float, relative: bool = False) -> Self: ...
    def arcTo(self, x: float, y: float, degrees: float, clockwise: bool, relative: bool = False) -> Self: ...
    def close(self) -> Self: ...
    def circle(self, radius: float) -> Self: ...

\end{lstlisting}
\end{minipage}
\end{figure*}

%% file: SupplementaryMaterial/Sections/details_about_code_format.tex
\section{CAD Code Representation Details}
Our data is processed based on the DeepCAD data format. For \textit{curve} parameters, we adopt the original parameter definitions used in DeepCAD. For example, the parameters for an \textit{arc} include the end-point coordinates $x$ and $y$, the sweep angle $\alpha$, and a counter-clockwise flag $f$ (also see Listing 1). The original DeepCAD representation does not include \textit{face} primitives; it only contains \textit{profile}. We obtain each \textit{face} by merging adjacent profiles.

For \textit{sketches}, unlike the representation used in DeepCAD, we define a sketch by specifying its associated plane (characterized by its \textit{origin}, \textit{normal}, and \textit{x-axis}) and the one or more \textit{faces} it contains. Regarding \textit{extrude} operations, the original DeepCAD data specifies an extrusion type $u$, which falls into one of three configurations: \textit{one-sided}, \textit{symmetric}, or \textit{two-sided}. We discard this parameter and instead use a unified representation \texttt{distance: Tuple[float, float]} to encode the two extrusion directions.

Furthermore, we quantize all continuous parameters into 256 levels and represent them using 8-bit integers. For angular values, we convert all continuous radian representations into discrete degree representations; for example, values in the range $[0, \pi]$ are mapped to $[0, 180]$.


%% file: SupplementaryMaterial/Sections/prompts_used_in_PLMs.tex
\section{Implementation Details of Baselines}
For Text2CAD \cite{text2cad}, CAD-Translator \cite{cad-translator}, CAD-LLaMA \cite{cad-llama}, and CAD-Coder \cite{cad-coder-sft}, we follow the original settings described in their respective papers to generate CAD models either through prompting or autoregressive decoding. The resulting CAD models are then uniformly converted into the JSON format defined by DeepCAD \cite{deepcad} for subsequent evaluation.

For baseline of pretrained large models (PLMs) such as GPT-4o \cite{gpt-4o}, we use the following prompts for the text-to-CAD and image-to-CAD tasks, respectively:

\ 

\textbf{Prompt for Text-to-CAD Task}: \textit{You are given a textual description of a CAD model.\textbackslash n\textbackslash nYour task is:\textbackslash n- Generate Python code using the CADLib API to implement the described CAD model.\textbackslash n- When using classes in CADLib, only use methods listed in the official CADLib API shown below. Do NOT assume or fabricate any methods.\textbackslash n- You may use standard Python libraries (e.g., math). Remember to \texttt{`}import\texttt{`} them if used.\textbackslash n\textbackslash nBelow is the official CADLib API for your reference:\textbackslash n\texttt{```}python\textbackslash n\{API Documentation\}\textbackslash n\texttt{```}Below is two examples:\textbackslash n\{Examples\}\textbackslash nYour final response MUST be Python code ONLY, using the following template:\textbackslash n\textbackslash n\texttt{```}python\textbackslash nfrom CADLib import Loop, Face, Sketch, Extrude, CADModel\textbackslash n\textbackslash ncad\_model = CADModel()\textbackslash n\# Add sketch and extrude to define the CAD model, and MUST use the variable name \texttt{`}cad\_model\texttt{`}\textbackslash n\texttt{```}\textbackslash n\textbackslash nNow generate the Python code for the following textual description: \{Input Description\}}

\textbf{Prompt for Image-to-CAD Task}: \textit{You are given the image of a CAD model: \textless image\textgreater\textbackslash n\textbackslash nYour task is:\textbackslash n- Generate Python code using the CADLib API to replicate the geometry shown in the image.\textbackslash n- When using classes in CADLib, only use methods listed in the official CADLib API shown below. Do NOT assume or fabricate any methods.\textbackslash n- You may use standard Python libraries (e.g., math). Remember to \texttt{`}import\texttt{`} them if used.\textbackslash n\textbackslash nBelow is the official CADLib API for your reference:\textbackslash n\texttt{```}python\textbackslash n\{API Documentation\}\textbackslash n\texttt{```}Below is two examples:\textbackslash n\{Examples\}\textbackslash nYour final response MUST be Python code ONLY, using the following template:\textbackslash n\textbackslash n\texttt{```}python\textbackslash nfrom CADLib import Loop, Face, Sketch, Extrude, CADModel\textbackslash n\textbackslash ncad\_model = CADModel()\textbackslash n\# Add sketch and extrude to define the CAD model, and MUST use the variable name \texttt{`}cad\_model\texttt{`}\textbackslash n\texttt{```}\textbackslash n\textbackslash nNow generate the Python code for the CAD model shown in the image.}

where \textit{\{API Documentation\}} refers to the content of Listing 1 along with explanations of the code, \textit{\{Examples\}} includes two examples of input-output pairs for the corresponding tasks, and \textit{\{Input Description\}} denotes the user-provided description in the text-to-CAD task.

%% file: SupplementaryMaterial/Sections/more_qualit_result.tex
\section{Additional Qualitative Results}
\subsubsection{More CAD Generation Cases}
\begin{figure*}[h]
    \centering
    \includegraphics[width=0.96\linewidth]{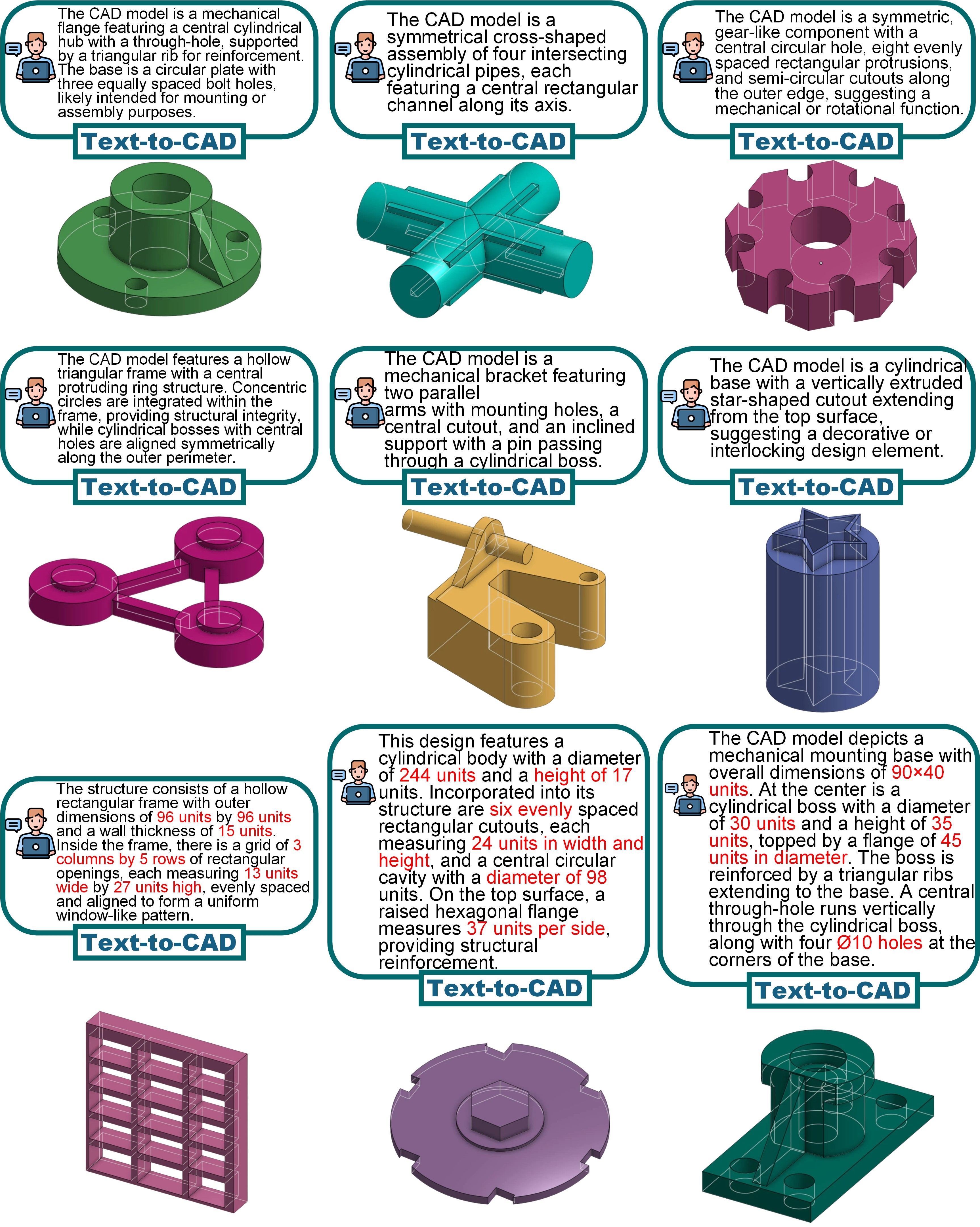}
    \caption{Visual results of ReCAD-VL on the text-to-CAD task.}
    \label{fig:figure_gen1}
\end{figure*}

\begin{figure*}[h]
    \centering
    \includegraphics[width=0.96\linewidth]{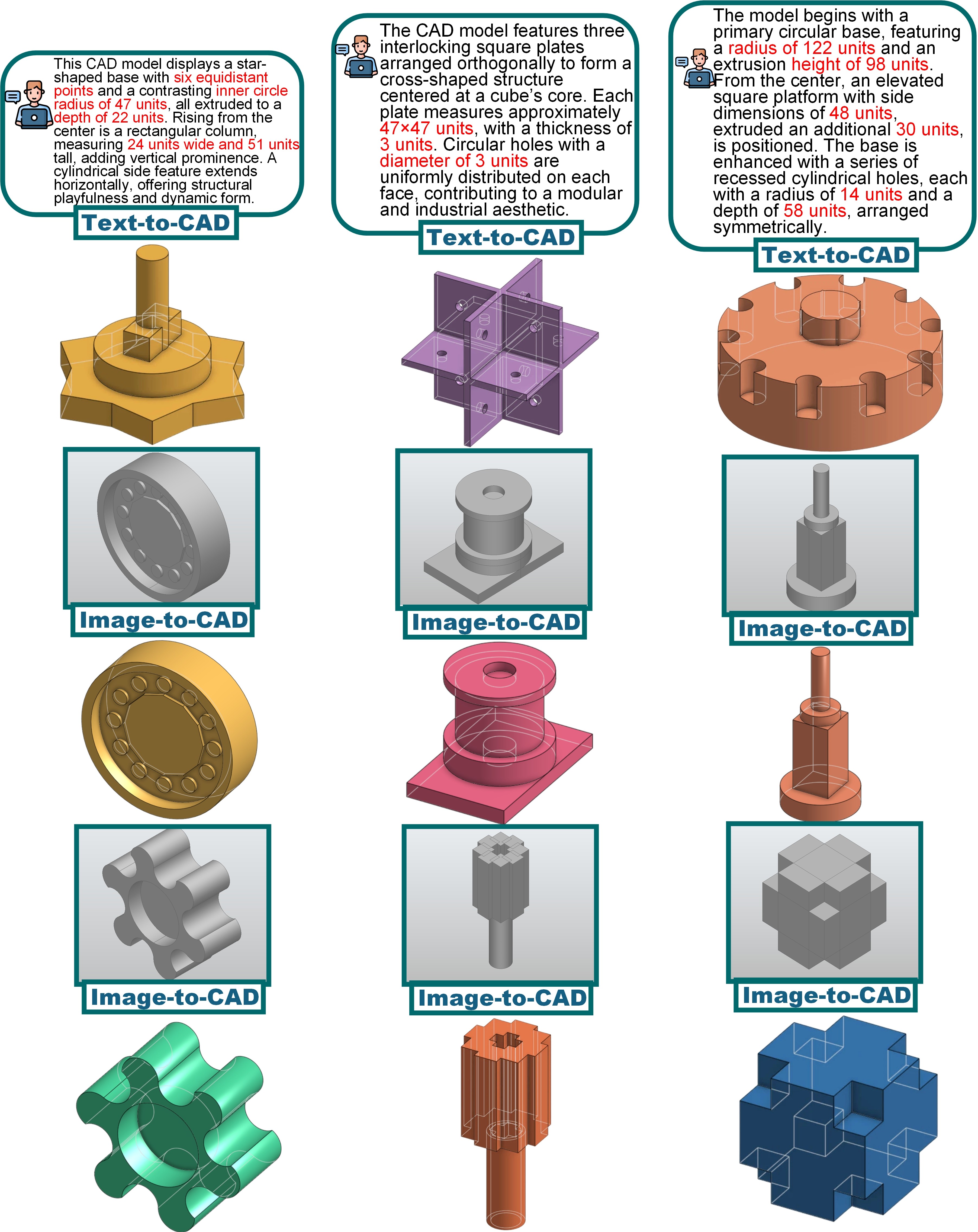}
    \caption{Visual results of ReCAD-VL on the text-to-CAD and image-to-CAD tasks.}
    \label{fig:figure_gen2}
\end{figure*}

We provide additional visualization results in Figures \ref{fig:figure_gen1} and \ref{fig:figure_gen2}, including examples from the text-to-CAD task, where CAD models are generated based on both abstract and precise textual descriptions, and the image-to-CAD task.
\subsubsection{More Zero-Shot Editing Cases}
\begin{figure*}[h]
    \centering
    \includegraphics[width=0.94\linewidth]{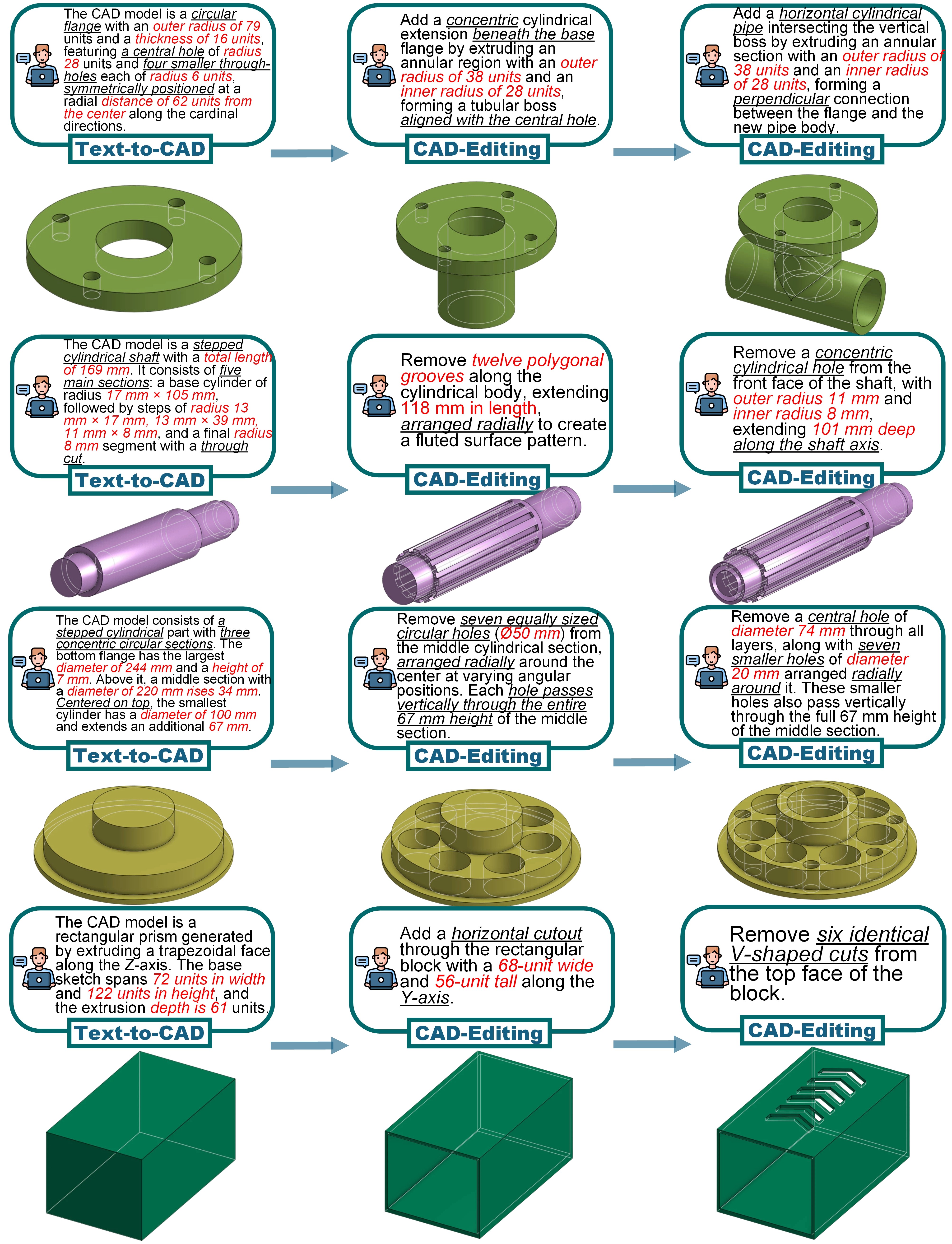}
    \caption{Visual results of ReCAD-VL on the zero-Shot editing task.}
    \label{fig:figure_edit}
\end{figure*}

Although ReCAD is not trained on CAD-specific tasks (such as CAD editing), it still exhibits impressive performance on them. Additional visual results demonstrating ReCAD’s capabilities in CAD editing are presented in Figure \ref{fig:figure_edit}.
\subsubsection{Full Model I/O Examples}
\begin{figure*}[h]
    \centering
    \includegraphics[width=0.90\linewidth]{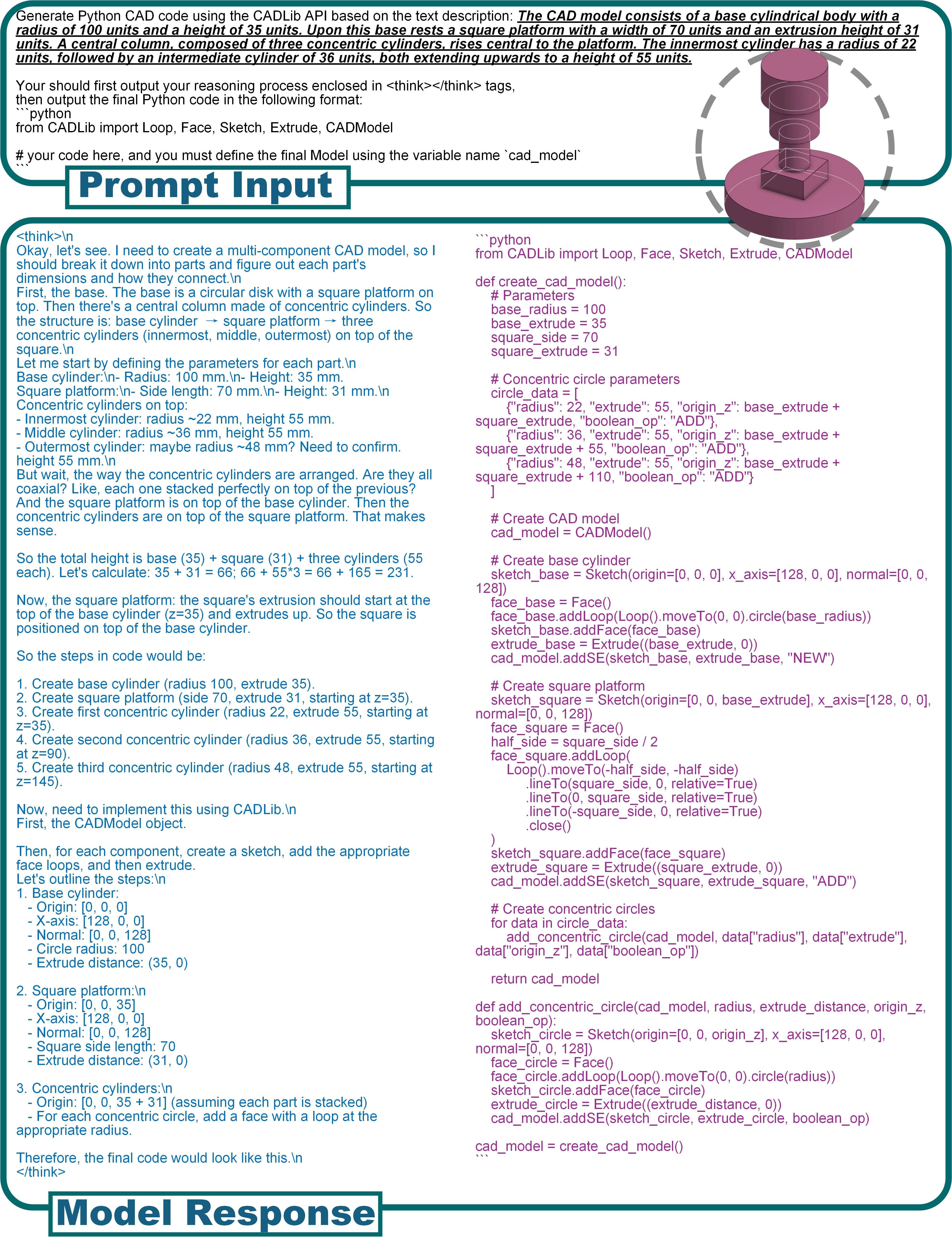}
    \caption{An input-output example for the text-to-CAD task using our proposed ReCAD-VL model. The bottom left section shows the Chain-of-Thought (CoT) reasoning content generated by the model, while the bottom right presents the final CAD code produced by ReCAD-VL. This example illustrates the model's step-by-step reasoning and generation process.}
    \label{fig:figure_io}
\end{figure*}

To clearly illustrate the step-by-step generation process of our model, we provide an input-output example for the text-to-CAD task, including the prompt used and the model's response. As shown in Figure~\ref{fig:figure_io}, the bottom left shows the model's Chain-of-Thought (CoT) reasoning content, while the bottom right shows the final CAD code generated by the model.

%% file: SupplementaryMaterial/Sections/failure_cases.tex
\section{Failure Cases}
\begin{figure*}[h]
    \centering
    \includegraphics[width=\linewidth]{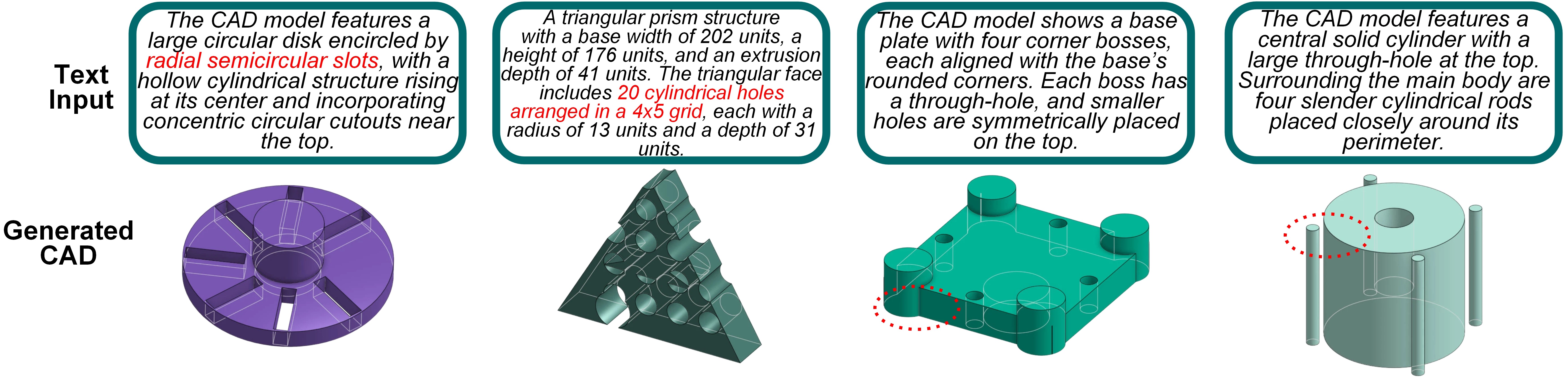}
    \caption{We categorize the failure cases into two types: those that do not match the input description (the two on the left), and those related to spatial parameterization errors (the two on the right).}
    \label{fig:failure-cases}
\end{figure*}

We analyze representative failure cases produced by our approach. As shown in Figure \ref{fig:failure-cases}, these cases can be broadly categorized into two types: mismatches with the input description (the two on the left), and errors related to spatial parameterization (the two on the right).

